\pdfoutput=1

\documentclass[11pt]{article}

\usepackage{acl}

\usepackage{times}
\usepackage{latexsym}

\usepackage[T1]{fontenc}

\usepackage[utf8]{inputenc}
\usepackage{graphicx} 
\usepackage{caption} 
\usepackage{amssymb}
\usepackage{amsmath}
\usepackage{multirow}

\usepackage{microtype}
\usepackage{subfigure}
\usepackage{geometry}
\usepackage[hang,flushmargin]{footmisc}

\title{{GMN}: Generative Multi-modal Network \\ for Practical Document Information Extraction}

\author{Haoyu Cao\textsuperscript{1}, Jiefeng Ma\textsuperscript{2}\thanks{~~Work is done during an internship at Tencent YouTu Lab.}~, Antai Guo\textsuperscript{1}, Yiqing Hu\textsuperscript{1}, Hao Liu\textsuperscript{1}, Deqiang Jiang\textsuperscript{1},\\ {\bf Yinsong Liu\textsuperscript{1}, Bo Ren\textsuperscript{1}} \\
        \textsuperscript{1}Tencent YouTu Lab \\
        \textsuperscript{2}University of Science and Technology of China \\
        \texttt{\{rechycao,ankerguo,hooverhu,ivanhliu,dqiangjiang\}@tencent.com} \\
        \texttt{jfma@mail.ustc.edu.cn}~~\texttt{\{jasonysliu,timren\}@tencent.com}
        }

\begin{document}
\maketitle
\begin{abstract}
    Document Information Extraction (DIE) has attracted increasing attention due to its various advanced applications in the real world.
    Although recent literature has already achieved competitive results, these approaches usually fail when dealing with complex documents with noisy OCR results or mutative layouts.
    This paper proposes Generative Multi-modal Network (GMN) for real-world scenarios to address these problems, which is a robust multi-modal generation method without predefined label categories. 
    With the carefully designed spatial encoder and modal-aware mask module, GMN can deal with complex documents that are hard to serialized into sequential order. Moreover, GMN tolerates errors in OCR results and requires no character-level annotation, which is vital because fine-grained annotation of numerous documents is laborious and even requires annotators with specialized domain knowledge.
    Extensive experiments show that GMN achieves new state-of-the-art performance on several public DIE datasets and surpasses other methods by a large margin, especially in realistic scenes.
  \end{abstract}

\section{Introduction}
Document Information Extraction (DIE) aims to map each document to a structured form consistent with the target ontology (\textit{e.g.}, database schema), which has recently become an increasingly important task. 
Recent research \cite{xu2020layoutlm,LayoutLMv2,VIE,zhang2020trie,StructLM} has achieved competitive results for information extraction in the idealized scenario with accurate OCR results, word-level annotations, and serialized document words in reading order.
These methods regard DIE as a Sequence Labeling (SL) task. Given OCR results of a document image, the traditional sequence labeling method first serializes words in reading order then classifies each input word into predefined categories.

\begin{figure}[t]
    \centering
\includegraphics[width=\linewidth]{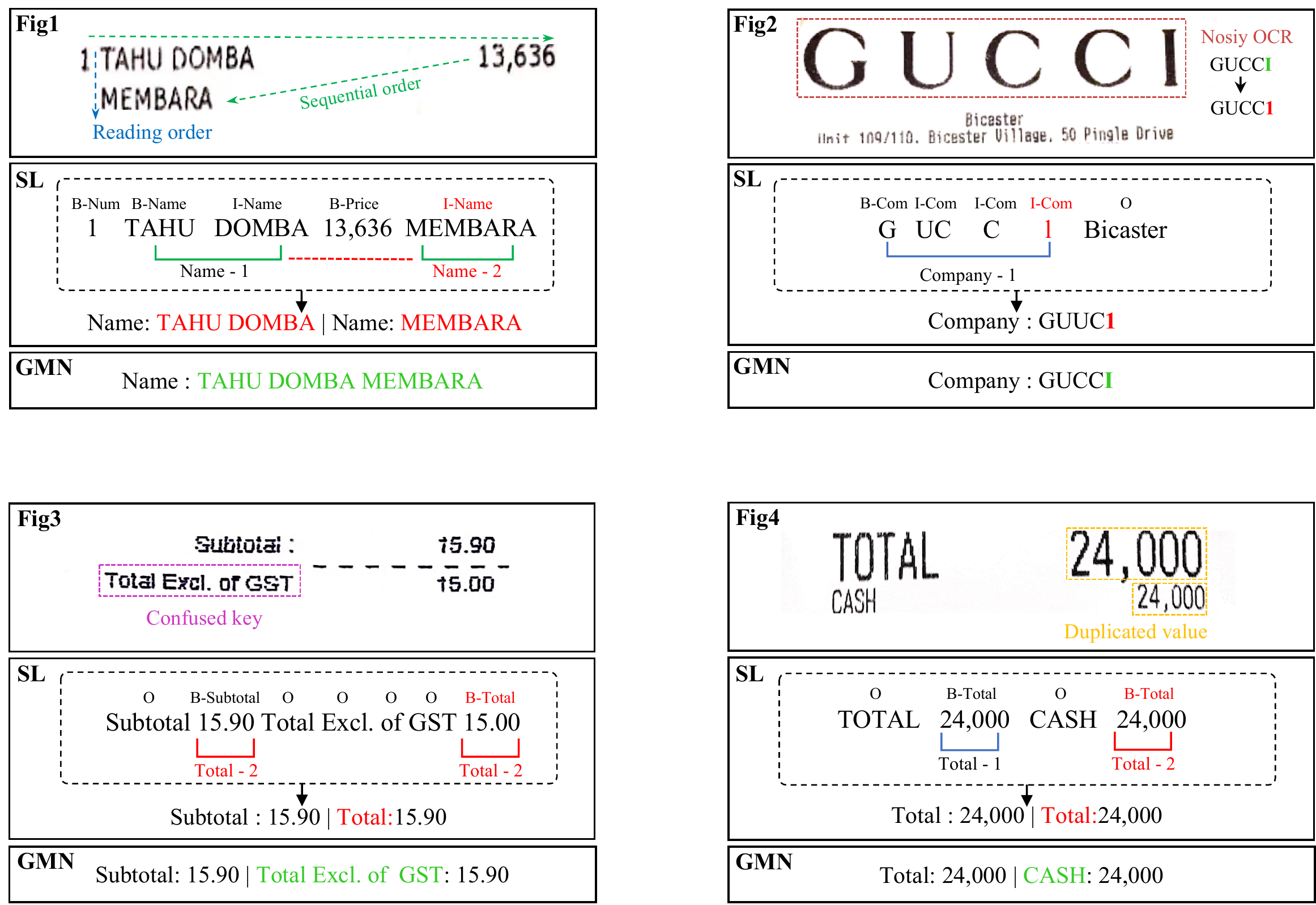}
    \caption{Examples in public DIE benchmarks with practical problems. The three rows from top to bottom in each sub-figure are 1) input images,  2) raw intermediate tags and final results generated by the SL method, and 3) results of our GMN method, respectively. The error parts are marked in red, while the correct parts are in green. Best viewed in color.}
    \label{fig:general_docs}
\end{figure}
  
  As shown in Figure~\ref{fig:general_docs}, multiple challenging problems for practical document understanding still exist in realistic scenes. 
  ~1)~Document serialization requires pre-composition processing, which is difficult in real scenarios with ambiguous word orders. One entity may be incorrectly divided into multiple entities when the input sequences are sorted by coordinates.
  ~2)~OCR results are usually noisy because of inevitable recognition errors.
  ~3)~The volume of keys in practical scenarios is generally substantial and expanded frequently. Existing sequence labeling methods could not identify undefined keys.
  ~4)~While facing duplicated values, collecting word-level annotations is necessary for sequence labeling methods. However, this is difficult in practical scenarios since they are costly and labor-intensive. 

  To address the limitations mentioned above, we propose a robust information extraction method named Generative Multi-modal Network (GMN) for practical document understanding.
  Unlike sequence labeling methods that label each input word with a predefined category, we regard DIE as a translation task that translates source OCR results to a structured format (like key-value pairs in this paper).
  We use UniLM \cite{unilm} as the basic model structure, which is a transformer-based pre-trained network that can handle both natural language understanding (NLU) and natural language generation (NLG) tasks simultaneously. Conditioned on a sequence of source words, GMN generates one word at each time step to compose a series of key-value pairs in them.
  
  Regarding the sequence serialization problem, a novel two-dimensional position embedding method is proposed while the original one-dimensional positional embedding in the transformer is removed because all information in document understanding can be acquired from 2D layouts. In this manner, GMN bypasses the serialization problem.
  Furthermore, benefiting from the large-scale self-supervised pre-training processed on a vast document collection, GMN can correct OCR errors commonly encountered in practical scenarios.
  Moreover, using a weakly supervised training strategy that utilizes only key information sequences as supervision, GMN needs no word-level annotations that are indispensable in traditional sequence labeling methods like LayoutLM \cite{xu2020layoutlm} and StructuralLM \cite{StructLM}.

  Experiments illustrate that the proposed GMN model outperforms several SOTA pre-trained models on benchmark datasets, including SROIE and CORD. The contributions of this paper are summarized as follows:
  \begin{itemize}
	\item [1)] We present GMN tailored for the DIE task, which is more applicable for practical scenarios, including lack of word-level annotations, OCR errors as well as various layouts.
	\item [2)] We propose a layout embedding method and multi-modal Transformer in a decoupling manner, which jointly models interactions between multiple modalities and avoids the reading order serialization. 
	\item [3)] Experiments on public DIE datasets demonstrate that the proposed method not only achieves a substantial performance improvement but also generalizes well to data under practical scenarios with unseen keys.
\end{itemize}

  \begin{figure*}[th]
    \centering
    \subfigure[The overall architecture of our generative multi-modal network, MD-Bert in encoder and decoder are with parameter sharing.]{
      \begin{minipage}[t]{1.0\linewidth}
      \centering
      \includegraphics[width=1.0\linewidth]{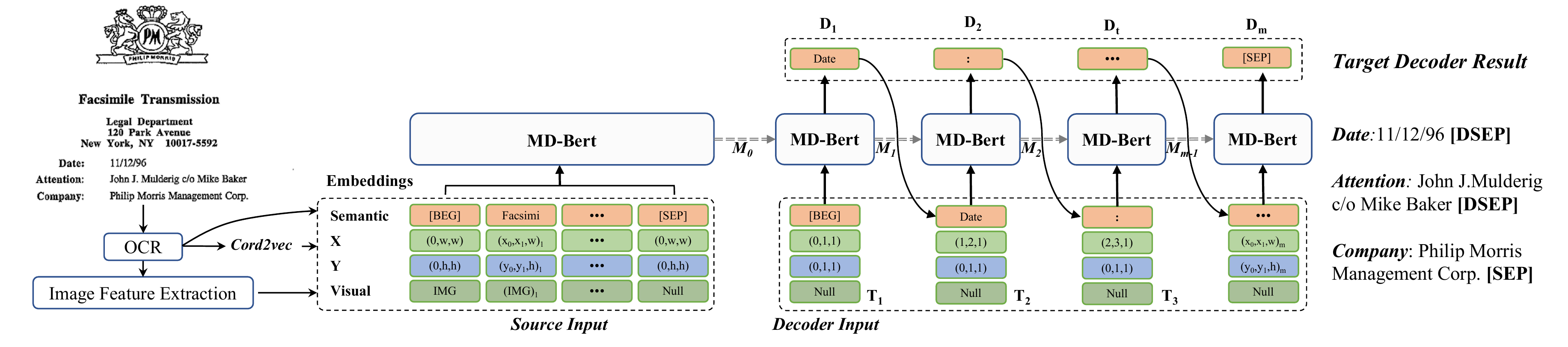}
      \label{fig:overall}
      \end{minipage}%
    }%
    \quad
    \subfigure[Detail of MD-Bert module, with internal memory updated recursively.]{
    \begin{minipage}[t]{1.0\linewidth}
    \centering
    \includegraphics[width=1.0\linewidth]{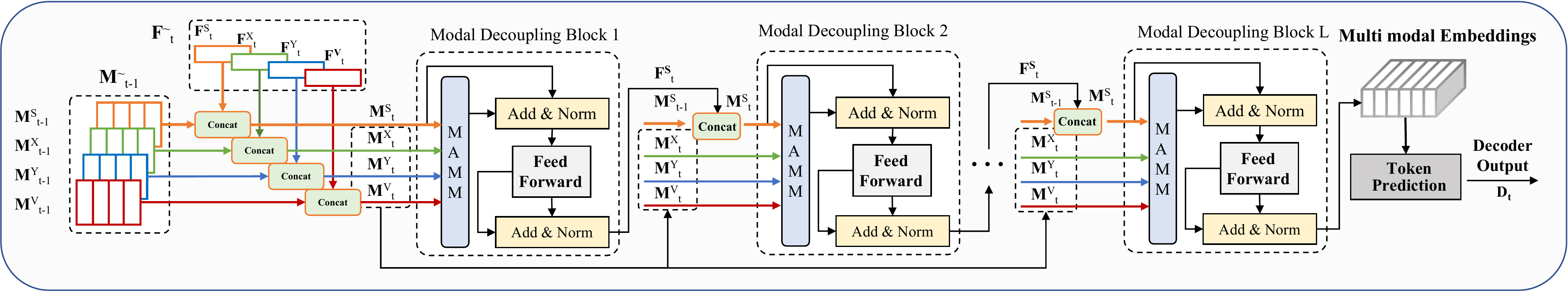}
    \label{fig:gmn}
    \end{minipage}
    }%
    \centering
    \caption{Architecture of our generative multi-modal network. 
    The MD-Bert module is composed of stacked multi-modal encoders, which fuses multi-modal features and iteratively generates structured results in a fixed order.}
    \label{fig:main_method}
  \end{figure*}
  
\section{RELATED WORKS}
  Traditional methods \cite{Tranditional1, Tranditional2, Tranditional3} on DIE tasks rely heavily on predefined rules, templates, and hand-crafted features, giving rise to difficulty in generalizing to unseen documents.
  With the development of deep learning technology, document information extraction methods have recently improved substantially in both performance and robustness.
  These deep learning-based methods can be classified into three categories: textual content-based methods, multi-modal-based methods, and pre-trained Transformer-based methods.

  {\bfseries Textual content-based methods}. \citet{palm2017cloudscan, sage2019recurrent} adopt the idea from natural language processing and use recurrent neural networks (RNN) to extract entities of interest from documents.
  However, they discard the layout information during the text serialization, which is crucial for document understanding.

  {\bfseries Multi-modal-based methods}.  Some works \cite{katti2018chargrid, hwang2020spatial} take the layout information into consideration and try to reconstruct character or word segmentation of the document.
  \citet{katti2018chargrid} encode each document page as a two-dimensional grid of characters that represents text representation with a two-dimensional layout.
  \citet{yu2021pick, majumder2020representation, zhang2020trie, wang2021tag} further integrate image embeddings for better feature extraction. 
  \citet{yu2021pick, MatchVIE} represent documents by graphs, with nodes representing word segments and edges either connecting all the nodes or only spatially near neighbors.
  Convolutional or recurrent mechanisms are then applied to the graph for predicting the field type of each node.
  However, due to the lack of large-scale pre-training, the robustness and accuracy of the model are relatively limited.

  {\bfseries Pre-trained Transformer-based methods}. 
  Recently, pre-trained models \cite{Bert, roberta} show effective knowledge transferability with large-scale training data and various self-supervised tasks.
  LayoutLM \cite{xu2020layoutlm} first proposes a document-level pre-training framework that semantic and layout information are jointly learned.
  LayoutLM V2 \cite{LayoutLMv2} further improves the LayoutLM model by integrating the image information in the pre-training stage.
  \citet{StructLM} propose the StructuralLM pre-training approach to exploit text block information.
  \textcolor{black}{
  Methods mentioned above all use one-dimensional position embeddings to model the word sequence, even with two-dimensional layout embeddings are involved, so that the reading order serialization in the document is required, which is challenging or even impossible due to the complex and diverse layout in the real world. What's more, they are all based on the classification of each input text segment to predefined labels, which means fine-grained annotations are indispensable and lack the ability to correct error OCR results.}

  On the contrary, the proposed GMN relies on a two-dimensional position embedding to bypass the serialization process and cross-modality encoders in a decoupling manner to model the layout information and the relative position of a word within a document simultaneously.

\section{METHODOLOGY}
In this section, we first introduce the overall architecture of GMN, followed by illustrating multi-modal feature extraction, generative pre-training model with multi-modal decoupling in detail, respectively.
\subsection{Overall Architecture}
\label{section:overall}
GMN aims at constructing an enhanced Transformer-based translator architecture for DIE for converting the document to structured, machine-readable data.
An overview of the architecture is as shown in Figure \ref{fig:main_method}. It mainly consists of two parts: the multi-modal feature extraction module and stacked cross-modality module named MD-Bert (Modal Decoupling Bert), which simultaneously serves as encoder and decoder following the design of UniLM.

The whole process can be summarized as 1) Multi-modality embeddings of source inputs are extracted through an advanced OCR engine and a small CNN; 2) The extracted features from different modalities are fused as ``multi-modal embeddings''  through MD-Bert along with memory updating for each layer at each time step;
3) Next, MD-Bert output the encoding results by applying token prediction on multi-modal embeddings; 
4) Finally, MD-Bert recursively generates structured results by taking multi-modal embeddings and accumulative memory as inputs until a terminator [SEP] is predicted.

\subsection{Multi-Modal Feature Extraction}
Based on the multi-modal information, including semantics, layout, and vision, we propose a unified layout embedding method named \textit{Cord2Vec} which simultaneously encodes sequence information and spatial information to avoid complex reading order serialization.
\subsubsection{ Semantic Embedding}
Intuitively, semantic contents are reliable signals to extract valuable information.
  The semantic content of each text fragment is acquired from the results of the OCR engine for practical application scenarios.
After text fragments are acquired and tokenized, the start indicator tag [BEG] is added in front of the input token sequence, and the end indicator tag [SEP] is also appended to the end. Extra padding tag [PAD] is used to unify the length of sequence with predefined batch length \begin{math} L \end{math}.
  In this way, we can get the input token sequence \begin{math} S \end{math} as
\begin{equation}
	\small
S = [[\text{BEG}], t_{1}, \cdots,t_{n}, [\text{SEP}], [\text{PAD}], \cdots ], |S| = L
\end{equation}

Here, \begin{math}
  t_{i}
\end{math} refers to \begin{math} i \end{math}-th token in OCR texts.
Moreover, though the sequence length of the input token sequence is fixed during training, GMN can handle variable lengths when making the inference due to the novel positional embedding method.

\subsubsection{Layout Embedding}

DIE task is a typical two-dimensional scene in which relative positions of words are essential evidence.
While the reading order serialization is challenging, we propose 
\textit{Cord2Vec}, a unified embedding method that fully utilizes spatial coordinates rather than one-dimensional sequence order information to bypass this problem.

As for the source input part, we normalize and discretize all coordinates to the integer in the range of \begin{math}
  [0, \alpha]
  \end{math}, here \begin{math}
    \alpha
  \end{math} is the max scale which is set to 1000 in our experiment.
  Then corner coordinates and edge lengths of each text fragment are gained using corresponding bounding boxes.
In order to enhance the tokens' interaction in the same box, two tuples \begin{math}(x_{0}, x_{1}, w), (y_{0}, y_{1}, h)\end{math} are used to represent the layout information. 
Here, \begin{math}
  (x_{0},y_{0}) 
\end{math} and \begin{math}
  (x_{1}, y_{1})
\end{math} are the top-left and bottom-right coordinates of each token, and 
\begin{math}
  w
\end{math} is the average width of tokens in the same box while \begin{math}
  h
\end{math} representing box height. Such embedding represents both the layout and the word order information.
As for target tokens generated by GMN which does not have the real coordinate, the \textit{Cord2Vec} assumes each token is tied in the grid of \begin{math}
  [W_{grid}, H_{grid}]
  \end{math} with row-first principle, and each token occupies a pixel with a width and height of 1.
After the layout information is acquired, we use two embedding layers to embed x-axis features and y-axis features separately as stated in Equation \ref{equation:PosEmb}.
\begin{equation} 
	\small
\begin{split}
&X_{i} = \operatorname{\textit{PosEmb2D}_{x}}(x_{0}, x_{1}, w),\\
&Y_{i} = \operatorname{\textit{PosEmb2D}_{y}} (y_{0}, y_{1}, h)
\end{split}
\label{equation:PosEmb}
\end{equation}
Here, \begin{math}
  \operatorname{\textit{PosEmb2D}_{x}}
\end{math} and \begin{math}
  \operatorname{\textit{PosEmb2D}_{y}}
\end{math} are the position embedding function which takes coordinate as input.
Each input element is embedded separately and then added together with an element-wise function.
Note that the placeholder such as [PAD] can be treated as some evenly divided grids, 
so their bounding box coordinates are easy to calculate.
An empty bounding box \begin{math} X_{\text{PAD}} = (0,0,0), Y_{\text{PAD}} = (0,0,0) \end{math} 
is attached to [PAD], and 
\begin{math} 
X_{\text{SEP}} = (0, w, w), Y_{\text{SEP}} = (0, h, h)
\end{math} 
is attached to other special tokens including [BEG] and [SEP].

\subsubsection{Visual Embedding}
We use ResNet-18 \cite{he2016deep} as the backbone of the visual encoder. 
Given a document page image \begin{math} I \end{math}, it is first resized to \begin{math} W * H \end{math} 
then fed into the visual backbone. After that, the feature map is scaled to a fixed size by average-pooling with the width being \begin{math} W/n \end{math} and height being \begin{math} H/n \end{math},
\textcolor{black}{ \begin{math} n \end{math} is the scaling scale. }
Finally, RoI Align \cite{he2017mask} is applied to extract each token's visual embedding with a fixed size.
The visual embedding of the \begin{math} i \end{math}-th token is denoted by \begin{math}
  v_{i} \in \left(v_{1}, v_{2}, v_{3}, \ldots, v_{L}\right)
\end{math}. 
For source input, visual embedding can be represented as
\begin{equation}
\small
    v_{i} = \operatorname{\textit{ROIAlign}}(\operatorname{\textit{ConvNet}}(Image), Pos_{i})
\end{equation}
Here, \begin{math} Pos_{i} \end{math} stands for the position of \begin{math} i \end{math}-th token, and
\begin{math}
  \operatorname{\textit{ConvNet}}
\end{math} 
is a convolutional neural network serving as feature extractor in terms of input image, and then \begin{math}
  \operatorname{\textit{ROIAlign}}
\end{math} takes the image feature and location as input, and extracts the corresponding image features. 
Note that the [BEG] token represents the full image feature,
and the other special tokens, as well as output tokens, are attached to the default null image feature.
 
\subsection{Generative Pre-training Model}
\subsubsection{Model Structure}
\begin{figure}[t]
\centering
  \includegraphics[width=0.4\textwidth]{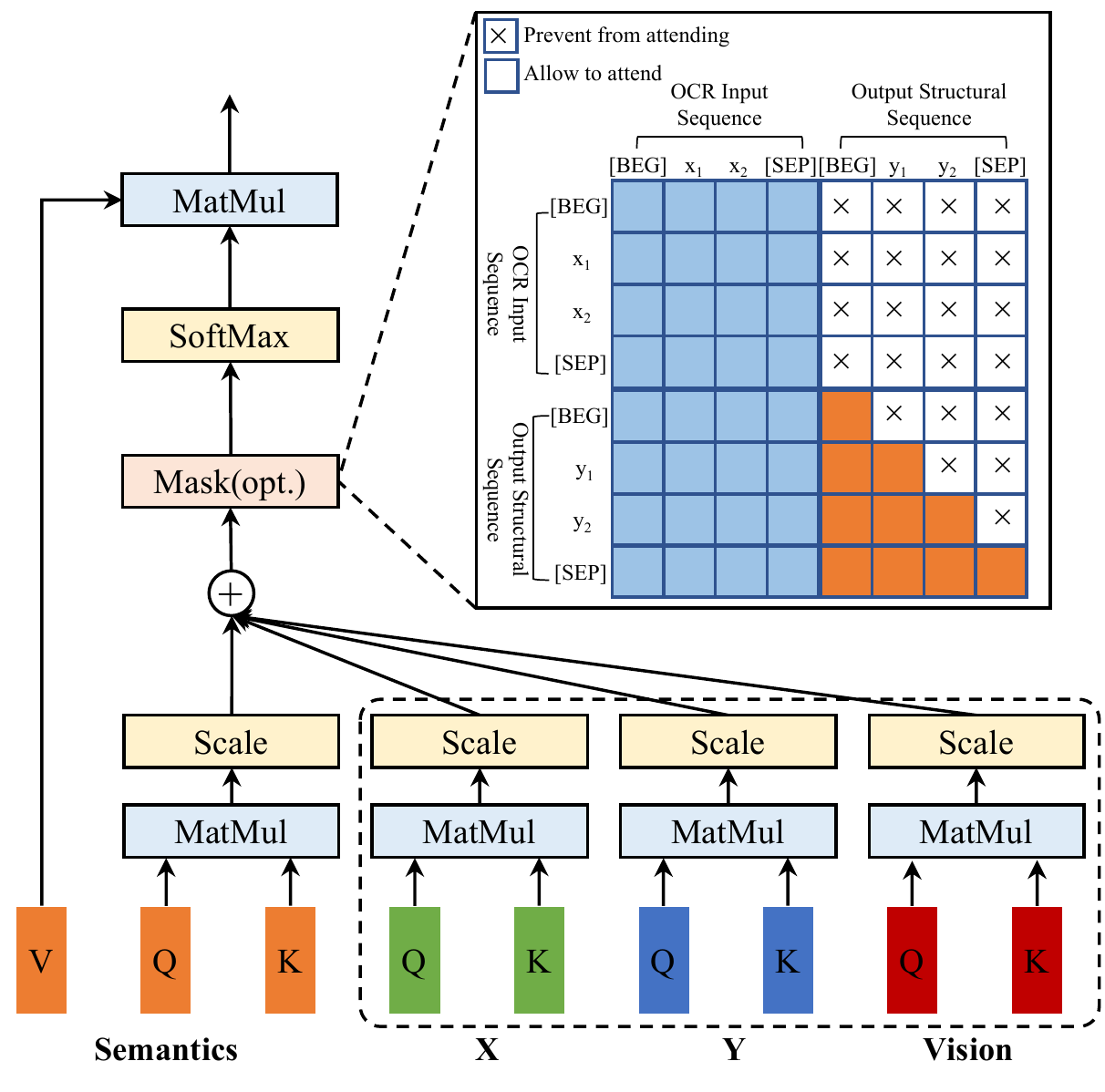}
  \caption{Modal-aware Mask Module: modalities fusion in decoupling manner. The sub-module for processing X are with parameter sharing for each layer, the same for Y and Visual sub-module.}
  \label{fig:MAMM}
\end{figure}

  In order to learn more general features and make full use of the pre-training data, we propose a unified encoder-decoder module named MD-Bert 
  which is composed of stacked hierarchical multi-modal Transformer encoders.
  The context of input tokens is from OCR result during the encoding stage, while already decoded tokens are also included in the decoding stage.
  To solve this problem, inspired by UniLM, we use masking to control which part of context the token should attend to when computing its contextualized representation,
  as shown in Figure \ref{fig:MAMM}. 
  The input features of semantics, layout and computer vision are mapped to hidden states by:
  \begin{math}
    \boldsymbol{F}_{0}^{S}=\left\{f_{1}^{S}, \ldots, f_{N}^{S}\right\}, 
    \boldsymbol{F}_{0}^{X}=\left\{f_{1}^{X}, \ldots, f_{N}^{X}\right\}, 
    \boldsymbol{F}_{0}^{Y}= \left\{f_{1}^{Y}, \ldots, f_{N}^{Y}\right\}, 
    \boldsymbol{F}_{0}^{V}= \left\{f_{1}^{V}, \ldots, f_{N}^{V}\right\}, 
  \end{math}
  a linear mapping is as follows:
\begin{equation}
  \small
  \begin{split}
   & f_{i}^{S}=\boldsymbol{W}_{S} x_{\text {i}}^{S}, f_{i}^{X}=\boldsymbol{W}_{X} x_{\text {i}}^{X}, \\
   & f_{i}^{Y}=\boldsymbol{W}_{Y} x_{\text {i}}^{Y}, f_{i}^{V}=\boldsymbol{W}_{V} x_{\text {i}}^{V}
   \label{eq:linear_mapping}
  \end{split}
\end{equation}
  where matrices \begin{math} \boldsymbol{W}_{S} \in \mathbb{R}^{d_{h} \times d_{\mathrm{S}}} \end{math},
  \begin{math} \boldsymbol{W}_{X} \in \mathbb{R}^{d_{h} \times d_{\mathrm{X}}} \end{math},
  \begin{math} \boldsymbol{W}_{Y} \in \mathbb{R}^{d_{h} \times d_{\mathrm{Y}}} \end{math},
  \begin{math} \boldsymbol{W}_{V} \in \mathbb{R}^{d_{h} \times d_{\mathrm{V}}} \end{math}
  are used to project features into hidden-state in \begin{math}
    d_{h} 
  \end{math} dimensions.
  MD-Bert takes \begin{math} F^{\sim} \end{math} and memory \begin{math} M^{\sim} \end{math} of history state as input, generates output and update memory \begin{math}
    M^{\sim}
  \end{math} step by step.
  where,
  \begin{math}
    {F}^{\sim} \in \{ F^{S}, F^{X}, F^{Y}, F^{V}\}, M^{\sim} \in \{ M^{S}, M^{X}, M^{Y}, M^{V}\}
  \end{math}, \begin{math}
    M^{\sim}
  \end{math} contains the history state of each layer and previous embeddings of the model. In the first timestep, \begin{math}
    M^{\sim}
  \end{math} is initialized from scratch, input \begin{math}
    T_{0}
  \end{math} means the full OCR result and MD-Bert acts as bi-directional encoder. For timestep  \begin{math}
    t
  \end{math}, where \begin{math}
    t \in [1, m]
  \end{math}, the model takes the output of the previous timestep or \begin{math} [BEG] \end{math} as input, and outputs the current result, MD-Bert acts as uni-directional decoder.
\subsubsection{Cross-Modality Encoder}
Traditional multi-modal models usually fuse different modal features by adding or concatenating them together, which inevitably introduces the unwanted correlations between each other during self attention procedure, \textit{e.g.} word-to-position, image-to-word.
However, these correlations are harmful to strengthening the model's capability as different data modalities are practically orthogonal, thus we need to design a customized pipeline for each one.
We propose MAMM (modal-aware mask module) encoder, a hierarchical structure multi-modal Transformer model in a decoupling manner that jointly models different modalities. 
As a consequence, the feature embedding decoupling is decomposed into three embeddings in GMN.
The MAMM module follows the design of a basic module in BERT, but replaces the multi-head attention with modal-aware multi-head attention.

It also contains feed-forward (FF) layers, residual connections, and layer normalizations (LN), meanwhile, parameters of modals are not shared.
When feeded with different modal content such as semantics, layout and computer vision, MAMM first calculates each modal's attention score separately, then added these attention scores together to get a fusion score, finally use this fusion score to apply masking and following operations on semantic content.
As shown in Figure \ref{fig:MAMM}, let \begin{math} \boldsymbol{F}^{\sim}_{l}=\left\{f^{\sim}_{1}, \ldots, f^{\sim}_{N}\right\} \end{math} be the encoded feature in the \begin{math}
  l
\end{math}-th layer. \begin{math} \boldsymbol{F}^{\sim}_{0} \end{math} is the vector of the input features as mentioned in Equation \ref{eq:linear_mapping}. Features output by the next layer \begin{math}
  \boldsymbol{F}^{\sim}_{l+1}
\end{math}
can be obtained via:
\begin{equation}
\small
  \boldsymbol{F}^{\sim}_{l-att}=\mathrm{\textit{LN}}\left(f_{\text {\textit{MAMM}}}\left(\boldsymbol{F}^{\sim}_{l}\right)+\boldsymbol{F}^{\sim}_{l}\right)
\end{equation}
\begin{equation}
\small
  \boldsymbol{F}^{\sim}_{l+1}=\mathrm{\textit{LN}}\left(f_{\mathrm{FF}}\left(\boldsymbol{F}^{\sim}_{l-att}\right)+\boldsymbol{F}^{\sim}_{l-att}\right)
\end{equation}
where \begin{math} f_{\textit{\text {MAMM }}}(\cdot) \end{math} is the modal-aware mask function defined as
\begin{equation}
\small
  f_{\text {MAMM }}\left(\boldsymbol{F}^{\sim}_{l}\right)=\operatorname{\textit{softmax}}\left( \operatorname{\textit{MaskProd}}\left(\boldsymbol{F}^{\sim}_{l} \right) \right) v\left(\boldsymbol{F}^{\sim}_{l}\right)
\end{equation}
\begin{equation}
\small
  \operatorname{\textit{MaskProd}}\left(\boldsymbol{F}^{\sim}_{l} \right) =
  \frac{
  q\left(\boldsymbol{F}^{\sim}_{l}\right)
  k\left(\boldsymbol{F}^{\sim}_{l}\right)^{\top}}
  {\sqrt[4]{d_{k}}}
  + f_{\text {\textit{Maskopt}}} 
\end{equation}
where \begin{math} q(\cdot) \end{math},  \begin{math} k(\cdot) \end{math}, \begin{math} v(\cdot) \end{math} are linear transformation layers applied to the proposals' feature, which represent the query, key and value in attention mechanism accordingly.
Benefited from the parameters' sharing among layers with regard to X, Y and Vision, GMN has comparable weights as Bert.
Symbol
\begin{math}
  d_{k}
\end{math} is the number of attention headers for normalization,
and the \begin{math} f_{\text {\textit{Maskopt}}} \end{math} is the Mask operation which controls the attention between each token.
In GMN, we apply full attention on all OCR input tokens, and input tokens of the model for  the output structural sequence can attend to the whole inputs as well as tokens that have been decoded which are like auto-regressive encoder.
Finally, \begin{math}
  \boldsymbol{F}^{\sim}_{l+1}
\end{math}  can be obtained by \begin{math}
  \boldsymbol{F}^{\sim}_{l-att}
\end{math}
via a feed-forward sub-layer composed of two fully-att connected layers of function \begin{math}
  f_{\mathrm{\textit{FF}}}(\cdot)
\end{math}. Hierarchically stacked layers form the multi-modal encoder.

\subsubsection{Pre-training Method}
Similar to UniLM, three cloze tasks including Unidirectional LM, Bidirectional LM and Sequence-to-Sequence LM are used in the GMN.
Meanwhile, we propose NER-LM for better entity correlation extraction.
The whole loss function is defined as,
\begin{equation}
\mathcal{L} = \mathcal{L}_{\text {uni-LM }}+\mathcal{L}_{\text {bi-LM}} +\mathcal{L}_{\text {s2s-LM}} + \mathcal{L}_{\text {NER-LM}}
\end{equation}

In a cloze task, we randomly choose some WordPiece \cite{wordpiece} tokens in the input, and replace them with the
special token \begin{math} [MASK] \end{math}. Then, we feed their corresponding output vectors computed by the Transformer
network into a softmax classifier to predict the masked token. The parameters of GMN are learned
to minimize cross-entropy loss, which is computed using the predicted tokens and the original tokens.

{\bfseries NER-LM} is an extension of sequence-to-sequence LM for better integrity constraints on the entity.
Given the source segment which includes entity values \begin{math} s_{1}, s_{2} \end{math}
, and the corresponding entity types \begin{math} n_{1}, n_{2} \end{math} as well as some background sentence \textit{e.g.} \begin{math} b_{1}, b_{2} \end{math},
we form the input format as A ``\begin{math} [BEG] s_{1} b_{1} s_{2} b_{2} [SEP] \end{math}'' and B ``\begin{math}
  [BEG] n_{1} s_{1} n_{2} s_{2} [SEP] \end{math}''.
Each token in A can access all others of A,
while each token in B can access all tokens of A as well as the preceded tokens in B.
The target entity in B is masked for prediction during training.

\section{EXPERIMENTS}
\subsection{Dataset}
\subsubsection{Pre-training Dataset}
Our model is pre-trained on the IIT-CDIP Test Collection 1.0 \cite{lewis2006building}, 
which contains more than 6 million documents, 
with more than 11 million scanned document images. 
Moreover, each document has its corresponding text and metadata stored in XML files which describe the properties of the document, such as the unique identity and document labels. 
And the NER-LM is pre-trained on the Enron Email Dataset \cite{klimt2004enron}, which contains 0.5 million emails generated by employees of the Enron Corporation. We follow the organization of the letter and generate the content on the image.
The structured information in the letters acts as an entity, such as subject, date, \textit{etc}.
\subsubsection{Fine-tuning Datasets}
We conduct experiments on three real-world public datasets, FUNSD-R, CORD and SROIE.

{\bfseries The FUNSD-R Dataset.} FUNSD \cite{FUNSD} is a public dataset of 199 fully annotated forms,
which is composed of 4 entity types (\textit{i.e.} Question, Answer, Header and Other). 
The original dataset has both semantic entity extraction (EE) and semantic entity linking (EL) tasks. 
It's noteworthy that the linking between different entities are complicated, one header entity may have linking to several question entities with more answer entities linked.

To better evaluate the system performance in the multi-key scenario, we relabel the dataset in key-value pairs format to tackle EE and EL tasks simultaneously. We named the new dataset FUNSD-R, which contains 1,421 keys for training and 397 keys for testing. Meanwhile, there are 267 keys in the test set that have not appeared in the training set.
FUNSD-R will be released soon.

{\bfseries The CORD Dataset.} The CORD \cite{cord} dataset contains 800 receipts for the training set,
100 for the validation set and 100 for the test set.
The dataset defines 30 fields under 4 categories and the task aims to label each word to the right field.

{\bfseries The SROIE Dataset.} SROIE \cite{SROIE} dataset contains 626 receipts for training and 347 receipts for testing.
Each entity of the receipt is annotated with pre-defined categories such as company, date, address, and total. 

To further investigate the capacity of our proposed method under more challenging scenarios, we expand ``SROIE'' and ``CORD'' datasets to  ``SROIE-S'' and ``CORD-S'' by shuffling the order of text lines and keep the box coordinates to simulate complex layouts. 
The evaluation metric is the exact match of the entity recognition results in the F1 score.
\subsection{Implementation Details}
{\bfseries Model Pre-training}.
We initialize the weight of GMN model with the pre-trained UniLM base model except for the position embedding layer and visual embedding layer. 
Specifically, our BASE model has the same architecture: a 12-layer Transformer with 768 hidden sizes, 
and 12 attention heads.
For the LARGE setting, our model has a 24-layer Transformer with 1,024 hidden sizes and 16 attention heads, which is initialized by the pre-trained UniLM LARGE model. 
For unidir-LM and bidir-LM methods, we select 15\% of the input tokens of sentence A for prediction. We replace these
masked tokens with the \begin{math} [MASK] \end{math} token 80\% of the time, a random token 10\% of the time, and an unchanged token 10\% of the time. 
For seq-to-seq LM and NER-LM, we select 15\% tokens of the sentence B. The target of the token is the next token.
Then, the model predicts the corresponding token with the cross-entropy loss.

In addition, we also add the two-dimensional position embedding and visual embedding for pre-training. 
Considering that the document layout may vary in different page sizes, 
we scale the actual coordinate to a ``virtual" coordinate: the actual coordinate is scaled to have a value from 0 to 1,000,
\textcolor{black}{and rescale the images to the size of $512\times512$.}

We train our model on 64 NVIDIA Tesla V100 32GB GPUs with a total batch size of 1,024. 
The Adam optimizer is used with an initial learning rate of 5e-5 and a linear decay learning rate schedule.

{\bfseries Task-specific Fine-tunings}.
We evaluate the model following the typical fine-tuning strategy and update all parameters in an end-to-end way on task-specific datasets.
We arrange the source OCR result from top to bottom and left to right. In addition, we add the ``\begin{math} [DSEP] \end{math}'' as the separator between text detection boxes.
In SROIE and CORD datasets, we construct the target key-value pairs in a certain order due to the keys being limited. (\textit{i.e.} company, date, address, total).
In the FUNSD dataset, we organize the target key-value pairs from top to bottom and left to right. We add the ``:'' as the separator between key and value and ``\begin{math} [DSEP] \end{math}'' as the separator between key-value pairs.
The max source length parameter is set to 768 in the SROIE and CORD datasets and 1536 in the FUNSD-R datasets, so input sequences below max length will be padding to the same length.
The model is trained for 100 epochs with a batch size of 48 and a learning rate of 5e-5.
Note that, the annotations of all GMN results are the weakly-supervised label of sentence-level while other methods use word-level annotations.

  \subsection{Comparison to State-of-the-Arts}
  We compare our method with several state-of-the-arts 
  on the FUNSD-R, SROIE and CORD benchmarks.
  We use the publicly available PyTorch models for BERT, UniLM and LayoutLM in all the experiment settings.
  The results of PICK \cite{yu2021pick}, MatchVIE \cite{MatchVIE}, BROS \cite{hong2020bros}, StrucTexT \cite{StrucTexT}, SPADE \cite{hwang2020spatial} and DocFormer \cite{DocFormer} are obtained from the original papers.
  
  \begin{table}
	\centering
	\small
    \begin{tabular} {lccc}
	\hline
      \bfseries \text { Model } & \bfseries \text { Precision } & \bfseries \text { Recall } & \bfseries \text { F1 } \\
      \hline
      Bert$_{\text{\textit{LARGE}}}$  & 0 & 0 & 0 \\
      LayoutLM$_{\text{\textit{LARGE}}}$  & 0 & 0 & 0 \\
      \hline
      GMN $_ {\text{\textit{BASE}}} $ & 0.5264 & 0.4866 & 0.5057 \\
      GMN $_ {\text{\textit{LARGE}}}$ & \bfseries 0.5568 & \bfseries 0.5116 & \bfseries 0.5333 \\
	\hline
    \end{tabular}
    \caption{Model results on the FUNSD-R dataset, methods based on sequence labeling yield under scene with larger amount of keys.}
    \label{tab:FUNSD-R}
  \end{table}

  { \bfseries Results under scene with larger amount of keys.}
  Table \ref{tab:FUNSD-R} shows the model results on the FUNSD-R dataset which is evaluated using entity-level precision, 
  recall and F1 score. 
  In the case of a large number of key categories, especially in the case 
  that some categories have not appeared in the training set, the method based on sequence labeling yield, 
  neither the Bert model, which only contains text modality nor the LayoutLM which also contains layout and visual modalities.
    The best performance is achieved by the \begin{math}
      GMN_{LARGE}
    \end{math}, where a significant improvement is observed compared to other methods.
    Note that, 67.25\% of keys have not appeared in the training set,
    This illustrates that the generative method in GMN is suitable for scenes with a large number of keys.

{\bfseries Results with Ground Truth Setting.}
Under this setting, the ground truth texts are adopted as model input.
As shown in Table \ref{tab:SROIE_GT} and Table \ref{tab:CORD_GT}, 
even using weakly supervised labels, our approach shows excellent performance on both SROIE and CORD, and yields new SOTA results,
which indicates that GMN has a powerful representational capability and can significantly boost the performance on DIE tasks.
\begin{table}
	\centering
	\small
  \begin{tabular}{lccc}
    \hline
    \bfseries \text { Model } & \bfseries \text { Precision } & \bfseries \text { Recall } & \bfseries \text { F1 } \\
    \hline $\text { BERT}_{\text{\textit{BASE}}}$  &0.9099&0.9099&0.9099\\
    $\text { UniLM}_{\text{\textit{BASE}}}$  &0.9459&0.9459&0.9459\\
    \hline
    $\text { BERT}_{\text{\textit{LARGE}}}$  &0.92&0.92&0.92\\
    $\text { UniLM}_{\text{\textit{LARGE}}}$  &0.9488&0.9488&0.9488\\
    $\text { LayoutLM}_{\text{\textit{LARGE}}}$  &0.9524&0.9524&0.9524\\
    $\text { LayoutLMv2}_{\text{\textit{LARGE}}}$  &0.9904&0.9661&0.9781\\
    \hline
    $\text { PICK}$  &0.9679&0.9546&0.9612\\
    $\text { MatchVIE}$  &-&-&0.9657\\
    $\text { BROS}$  &-&-&0.9662\\
    $\text { StrucTexT}$  & 0.9584 & \bfseries 0.9852 & 0.9688\\
    \hline
    $\text { GMN} _ {\text{\textit{BASE}}} $ & 0.9853 & 0.9633 & 0.9741 \\
    $\text { GMN} _ {\text{\textit{LARGE}}} $ & \bfseries 0.9956 & 0.9690 & \bfseries 0.9821 \\
    \hline
  \end{tabular}
  \caption{Model results on the SROIE dataset with Ground Truth Setting.}
  \label{tab:SROIE_GT}
\end{table}

\begin{table}
	\centering
	\small
  \begin{tabular}{lccc}
    \hline
    \bfseries \text { Model } & \bfseries \text { Precision } & \bfseries \text { Recall } & \bfseries \text { F1 } \\
    \hline $\text { BERT}_{\text{\textit{BASE}}}$  &0.8833&0.9107&0.8968\\
    $\text { UniLM}_{\text{\textit{BASE}}}$  &0.8987&0.9198&0.9092\\
    \hline
    $\text { BERT}_{\text{\textit{LARGE}}}$ &0.8886&0.9168&0.9025\\
    $\text { UniLM}_{\text{\textit{LARGE}}}$  &0.9123&0.9289&0.9205\\
    $\text { LayoutLM}_{\text{\textit{LARGE}}}$ &0.9432&0.9554&0.9493\\
    $\text { LayoutLMv2}_{\text{\textit{LARGE}}}$ &0.9565&0.9637&0.9601\\
    $\text { SPADE}$   &-&-&0.925\\
    $\text { DocFormer}$   &\bfseries 0.9725&0.9674&0.9699\\
    $\text { BROS}$   &-&-&0.9728\\
    \hline
     $\text { GMN} _ {\text{\textit{BASE}}} $ & 0.9547 & 0.9576 & 0.9562 \\
     $\text { GMN} _ {\text{\textit{LARGE}}} $ & 0.9693 & \bfseries 0.9798 & \bfseries 0.9745 \\
	 \hline
  \end{tabular}
  \caption{Model results on the CORD dataset with Ground Truth Setting.}
  \label{tab:CORD_GT}
\end{table}

{\bfseries Results with End-to-End Setting.}
 We adopt Tesseract as OCR engine to get the OCR result of public datasets. 
 It's worth noting that there are exist some OCR errors, the sequence labeling method can not handle, 
 but in our GMN, the matching process between OCR results and ground truth is avoided thanks to the novel layout embedding method \textcolor{black}{in} an end-to-end training setting.
 The performances are shown in Table \ref{tab:sroie_cord_E2E}. Our method shows new state-of-the-art performance benefits from the ability to error correction.
 A detailed analysis of it is introduced in case studies \ref{sec:appendix_case}.
 
\begin{table}[]
\centering
\small
\begin{tabular}{lccc}
\hline
\multicolumn{1}{c}{\multirow{2}{*}{\bfseries \text {Model}}}  &\multicolumn{3}{c}{\bfseries \text {SROIE-E2E}} \\ \cline{2-4}
\multicolumn{1}{c}{}    & \bfseries \text {Precision} & \bfseries \text {Recall} & \bfseries \text {F1}\\
\hline
$\text { BERT}_{\text{\textit{LARGE}}}$     & 0.4066&0.3876&0.3969      \\
$\text { LayoutLM}_{\text{\textit{LARGE}}}$ & 0.4414&0.4236&0.4323      \\ \hline
$\text { GMN} _ {\text{\textit{BASE}}} $    & 0.7324 & 0.7161 & 0.7242      \\
$\text { GMN} _ {\text{\textit{LARGE}}} $   & \bfseries 0.7543 & \bfseries 0.7334 & \bfseries 0.7437      \\ \hline
\multicolumn{1}{c}{\multirow{2}{*}{\bfseries \text {Model}}}  &\multicolumn{3}{c}{\bfseries \text {CORD-E2E}}  \\ \cline{2-4}
\multicolumn{1}{c}{}    & \bfseries \text {Precision} & \bfseries \text {Recall} & \bfseries \text {F1}\\
\hline
$\text { BERT}_{\text{\textit{LARGE}}}$     & 0.6313 & 0.6724 & 0.6512      \\
$\text { LayoutLM}_{\text{\textit{LARGE}}}$ & 0.6684 & 0.7086 & 0.6879      \\ \hline
$\text { GMN} _ {\text{\textit{BASE}}} $    & 0.7840 & 0.8133 & 0.7984      \\
$\text { GMN} _ {\text{\textit{LARGE}}} $   & \bfseries 0.8165 & \bfseries 0.8368 & \bfseries 0.8265      \\ \hline
\end{tabular}
\caption{Model results on the SROIE and CORD datasets with End-to-End Setting.}
\label{tab:sroie_cord_E2E}
\end{table}
 
{\bfseries Results with Position Shuffle Setting.}
In order to verify the robustness of our two-dimensional embedding method, we apply a shuffling operation on boxes of the test dataset.
As shown in Table \ref{tab:sroie_cord_ps}, compared with models that have one-dimensional position embeddings, our method is more robust to input disruption with a big gap.

\begin{table}[]
\centering
\small
\begin{tabular}{lccc}
\hline
\multicolumn{1}{c}{\multirow{2}{*}{\bfseries \text {Model}}}  &\multicolumn{3}{c}{\bfseries \text {SROIE-S}} \\ \cline{2-4}
\multicolumn{1}{c}{}    & \bfseries \text {Precision} & \bfseries \text {Recall} & \bfseries \text {F1}\\
\hline
    $\text { BERT}_{\text{\textit{BASE}}}$ &0.0702&0.0490&0.0577\\
    $\text { LayoutLM}_{\text{\textit{BASE}}}$  &0.7169&0.6880&0.7022\\
    $\text { GMN} _ {\text{\textit{BASE}}} $ & \bfseries 0.9679 & \bfseries 0.9424 & \bfseries 0.9550 \\
\hline
\multicolumn{1}{c}{\multirow{2}{*}{\bfseries \text {Model}}}  &\multicolumn{3}{c}{\bfseries \text {CORD-S}}  \\ \cline{2-4}
\multicolumn{1}{c}{}    & \bfseries \text {Precision} & \bfseries \text {Recall} & \bfseries \text {F1}\\
\hline
    $\text { BERT}_{\text{\textit{BASE}}}$ & 0.1235 & 0.1384 & 0.1305\\
    $\text { LayoutLM}_{\text{\textit{BASE}}}$  & 0.6917 & 0.7139 & 0.7026\\
    $\text { GMN} _ {\text{\textit{BASE}}} $ & \bfseries 0.9345 & \bfseries 0.9488 & \bfseries 0.9416 \\
\hline
\end{tabular}
\caption{Model results on the SROIE-S and CORD-S datasets with Position Shuffle Setting.}
\label{tab:sroie_cord_ps}
\end{table}

\begin{table}
	\small
	\centering
  \begin{tabular}{lccc}
    \hline
    \bfseries \text { Model } & \bfseries \text { Precision } & \bfseries \text { Recall } & \bfseries \text { F1 } \\
    \hline
    GMN & \bfseries 0.9693 & \bfseries 0.9798 & \bfseries 0.9745 \\
    GMN w/o Image & 0.9623 & 0.9608 & 0.9616 \\
    GMN w/o NER-LM & 0.96 & 0.9585 & 0.9593 \\
    GMN w/o MAMM & 0.9576 & 0.9547 & 0.9562 \\
    \hline
\end{tabular}
\caption{Model results of different components of our method on the CORD dataset.}
\label{tab:Ablation}
\vspace{-1.0em}
\end{table}

\subsection{Ablation Study}
An ablation study is conducted to demonstrate the effects of different modules in the proposed model.
We remove some components to construct several comparable baselines on the CORD dataset.
The statistics are listed in Table \ref{tab:Ablation}. 

The ``GMN w/o MAMM'' means using the same multi-modal feature as LayoutLM. 
Compared with LayoutLM, MAMM brings about 1.82\% improvement of F1, which verifies the validation of MAMM.
The ``GMN w/o Image'' means removing the image feature extraction. Experiment results show that visual modality  can also improve the performance. 
Moreover, with NER-LM considered, the performance of information extraction increases to 97.45\%.
Extended experiments including attention visualization analysis and case studies can refer to Appendix \ref{sec:appendix_att} and \ref{sec:appendix_case}.

\section{CONCLUSION}
  In this work, we propose a Generative Multi-modal Network (GMN) for practical document information extraction.
  Since GMN is a generation method including no pre-defined label category, 
  it supports scenes that contain unknown similar keys and tolerates OCR errors, meanwhile requires no character-level annotation.
  We conduct extensive experiments on publicly available datasets to validate our method, experimental results demonstrate that GMN achieves state-of-the-art results on several public DIE datasets, especially in the practical scenarios. 
  Though our GMN significantly outperforms other DIE models, 
  there still exists potential to be exploited as regard to practical scenarios. 
  \textcolor{black}{In order to cope with complicated layout information as well as ambiguous semantic representations, we argue that more attention should be paid to the modality embedding and interaction strategy, which has more opportunity to handle such difficult cases.}
  
\bibliography{anthology,custom}
\bibliographystyle{acl_natbib}

\noindent \textbf{Appendix}
\appendix

\begin{figure*}[!t]
\centering
	\includegraphics[width=0.85\textwidth]{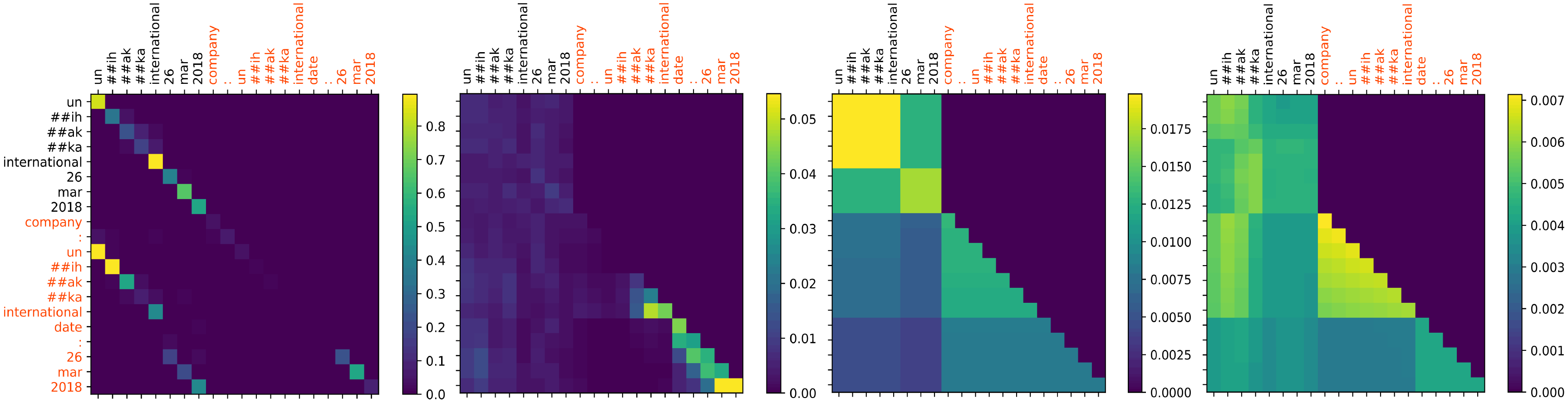}
	\caption{Visualization of the attention map of the multi-head Transformer, representing semantic/X-coord/Y-coord/visual attention results in a row respectively. The decoding result is marked in orange. Best viewed in color.}
	\label{fig:attention}
\end{figure*}

\begin{figure*}[!t]
\centering
	\includegraphics[width=0.86\textwidth]{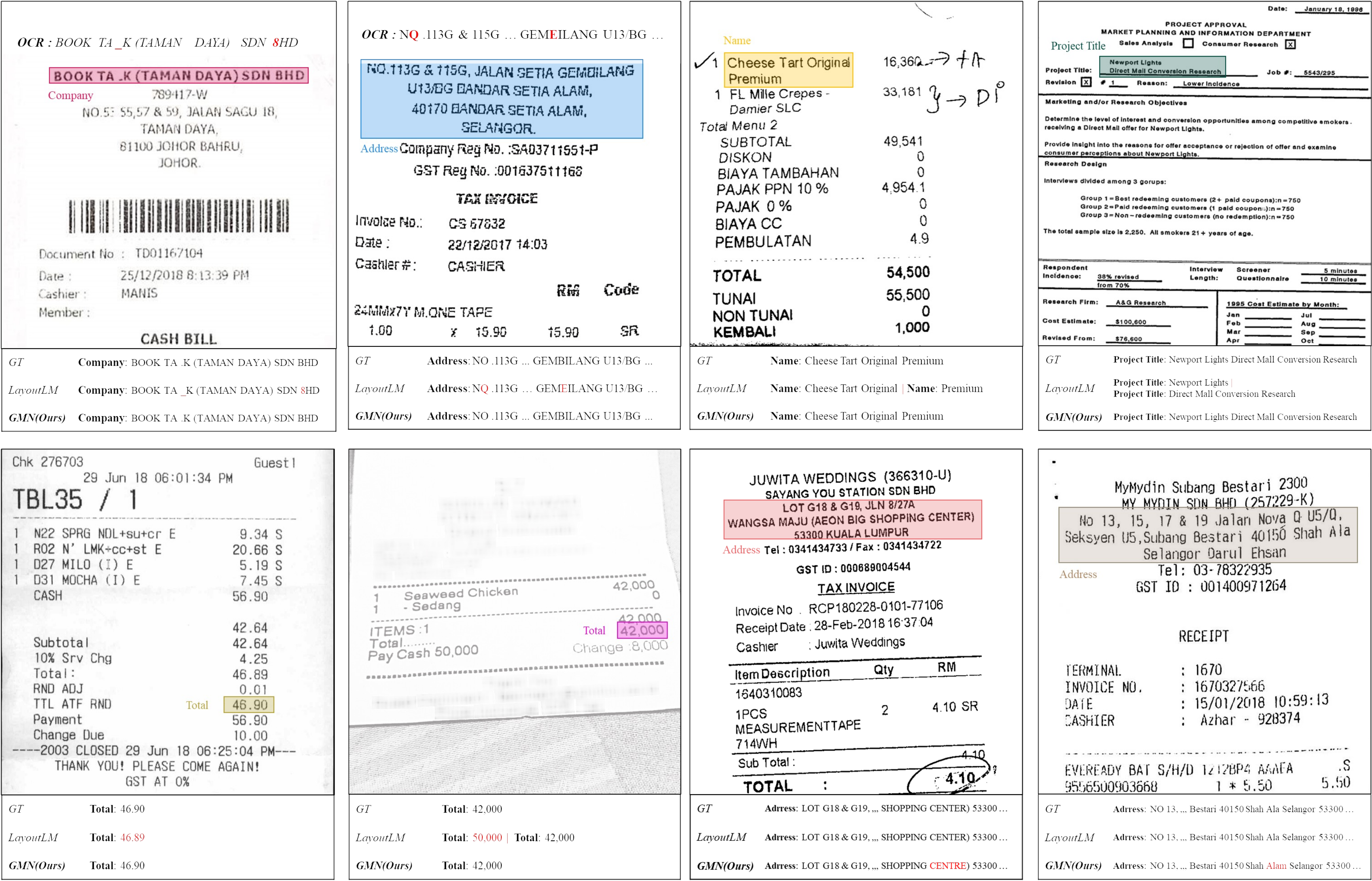}
	\caption{Samples from SROIE/CORD/FUNSD datasets, key examples are highlighted by color boxes. Best viewed in color.}
	\label{fig:show_case}
\end{figure*}

\section{Attention Visualization}

To further explore what context information is focused by our GMN, we visualize the attention map of the multi-head Transformer, as shown in Figure \ref{fig:attention}. The input tokens of the model are marked in black and the decoding results are marked in orange, while X-axis represents the attended tokens. 
As shown in Figure \ref{fig:MAMM}, we use the \textit{Mask} operator to control the attention between each token. The input OCR tokens can attend to each other but the output tokens can only be attended to the already decoded tokens. 
Consequently, the upper right area of the attention map has no active response, and the area in the lower right corner shows a stepped pattern.

We can observe that the semantic attention mechanism plays an important role in modeling local dependence.
In semantic attention, the input OCR tokens mainly focus on themselves and their nearby semantically relevant parts. 
In contrast, decoded tokens mostly focus on the counterparts in the original tokens, showing a reasonable alignment.
Meanwhile, layout and visual attention mechanisms focus on more global information, complementing the semantic attention mechanism.
\label{sec:appendix_att}

\section{Case studies}

The motivation behind GMN is to tackle the practical DIE tasks. To verify this, we show some examples of the output of LayoutLM and GMN,
as shown in Figure \ref{fig:show_case}. 
In the sub-figure A and B, GMN successfully corrects the recognition error of OCR results thanks to semantic learning on a large-scale corpus. 
In the sub-figure C\textasciitilde F, GMN accurately generates the key-value pairs with complex layouts and ambiguous contexts thanks to the novel position embedding method, in comparison LayoutLM is unable to merge the value entities correctly. 
It's noteworthy that the sub-figures G and H are failed cases, which are caused by semantic obfuscation and reasonable complement to missing character.
These examples show that GMN is capable of correcting OCR errors and predicting more accurately in practical scenarios.
\label{sec:appendix_case}

\end{document}